\documentclass[11pt,a4paper]{article}

\usepackage[utf8]{inputenc}

\usepackage[margin=2.5cm]{geometry}
\usepackage{amsmath,amssymb}
\usepackage{graphicx}
\usepackage{booktabs}
\usepackage{multirow}
\usepackage{array}
\usepackage[table]{xcolor}
\usepackage{caption}
\usepackage{subcaption}
\usepackage{microtype}
\usepackage{authblk}
\usepackage[numbers,sort&compress]{natbib}
\usepackage[hidelinks]{hyperref}
\usepackage[capitalise,noabbrev]{cleveref}
\usepackage{enumitem}

\setlist{itemsep=2pt,topsep=4pt}

\newcommand{\cog}{\texttt{cog}}
\newcommand{\vol}{\texttt{vol}}
\newcommand{\bio}{\texttt{bio}}
\newcommand{\pet}{\texttt{pet}}
\newcommand{\demo}{\texttt{demo}}
\newcommand{\ci}[2]{\,{\scriptsize[#1, #2]}}

\title{Anatomical Grounding and Leakage-Aware Multimodal Contrastive Learning for Alzheimer's Disease Classification from Structural MRI}

\author[1]{Paul-Gabriel Nicolae}
\author[1]{Irina Mocanu}
\author[ ]{for the Alzheimer's Disease Neuroimaging Initiative\thanks{Data used in preparation of this article were obtained from the Alzheimer's Disease Neuroimaging Initiative (ADNI) database (\url{adni.loni.usc.edu}). As such, the investigators within the ADNI contributed to the design and implementation of ADNI and/or provided data but did not participate in the analysis or writing of this report. A complete listing of ADNI investigators can be found at \url{http://adni.loni.usc.edu/wp-content/uploads/how_to_apply/ADNI_Acknowledgement_List.pdf}.}}
\affil[1]{Computer Science and Engineering Department, Faculty of Automatic Control and Computers, National University of Science and Technology POLITEHNICA Bucharest, Romania}
\date{}

\begin{document}
\maketitle

\begin{abstract}
Deep networks trained on structural MRI for Alzheimer's disease (AD) staging often reach reasonable accuracy while attending to anatomically irrelevant regions, and multimodal models that add clinical tables frequently rely on variables that were used to assign the diagnostic label in the first place. We study both issues with a deliberately lightweight slice-based encoder (ResNet18 with a one-layer Transformer over slices) on 1{,}075 baseline T1-weighted scans from ADNI-1. First, we use FastSurfer segmentations as an anatomical reference: YOLOv8 models trained on segmentation-derived labels localize Alzheimer-relevant structures with mAP$_{50}$ above 0.96, and a Grad-CAM comparison shows that the image-only classifier frequently attends to the skull, orbits and background. Second, we adapt a CLIP-style image--tabular contrastive framework and organize ADNIMERGE variables along a \emph{label-leakage spectrum}. Fusion with cognitive scores yields 87.3\% three-way accuracy, which we treat as a leakage-driven upper bound rather than an imaging result; fusion with regional volumes yields 73.0\%. We observe that the choice of contrastive target changes what the image encoder learns: on MCI vs.\ CN, the image-only head reaches 52.4\% when the encoder is aligned to cognitive scores and 73.8\% when aligned to volumes, although no tabular input is used at inference. Third, restricting the input to a per-subject crop of the medial temporal lobe raises image-only three-way accuracy from 58.7\% to 65.1\%. All results come from single runs on a small balanced test set, and we report confidence intervals and the protocol differences that prevent direct comparison with published numbers.
\end{abstract}


\section{Introduction}

Alzheimer's disease (AD) is the most common cause of dementia, and its prodromal stage, mild cognitive impairment (MCI), is the window in which intervention is expected to be most useful. Structural T1-weighted MRI is widely available and captures the atrophy that accompanies the disease, which has made it a standard input for automatic staging into cognitively normal (CN), MCI and AD. Separating AD from CN on MRI is comparatively easy, because late-stage atrophy is visible; separating MCI from CN is much harder, because baseline differences are subtle~\cite{wen2020reproducible}.

Two practical problems recur in this literature. The first concerns \emph{what the image model looks at}. A convolutional network that receives whole slices has no reason to restrict itself to brain tissue, and can exploit contrast at the skull, orbits or ventricle boundaries. The second concerns \emph{what multimodal models are given}. Large cohorts such as ADNI~\cite{jack2008adni} ship rich clinical tables, and recent image--tabular contrastive frameworks~\cite{huang2023multimodal,hager2023best} report strong gains when these tables are fused with MRI. However, several table columns, most notably cognitive scores such as the MMSE and CDR, are part of the criteria used to assign the ADNI diagnostic label~\cite{petersen2010adni}. High accuracy obtained with these inputs measures how well a model can recover the labelling rule, not how much diagnostic information the image contributes.

In this paper we examine both problems within one lightweight pipeline, a 2D ResNet18 slice encoder aggregated by a single Transformer layer, trained on ADNI-1 baseline scans. We make the following contributions:
\begin{enumerate}
    \item \textbf{Anatomical reference and diagnosis of the image model.} We use FastSurfer~\cite{henschel2020fastsurfer} segmentations to derive slice ranges and detection/segmentation labels for Alzheimer-relevant structures, train YOLOv8 models on them, and compare their outputs with Grad-CAM maps of the image classifier (\cref{sec:anatomy}).
    \item \textbf{A leakage-aware multimodal protocol.} We adapt CLIP-style contrastive fusion~\cite{radford2021clip,huang2023multimodal} to our encoder and evaluate ADNIMERGE modality groups separately, ordered by how directly they encode the label (\cref{sec:multimodal}).
    \item \textbf{The contrastive target shapes the image encoder.} Our main finding: with architecture, split and training held fixed, aligning the encoder to anatomical volumes instead of cognitive scores produces an image-only representation that is 21 points better on MCI vs.\ CN, even though no tabular input is used at inference (\cref{sec:target}).
    \item \textbf{Anatomically guided input.} A per-subject crop of the medial temporal lobe (MTL), derived from the same segmentations, improves image-only accuracy without changing the encoder (\cref{sec:mtl}).
\end{enumerate}
We also make explicit why our numbers should not be read as outperforming prior work: the test set is small (63 scans), results are single-run, and the evaluation protocol differs from published ones in several respects (\cref{sec:discussion}).

\section{Related Work}

\paragraph{MRI-based AD classification.} Convolutional networks on 2D slices or 3D volumes are the dominant approach. 3D ResNets~\cite{ebrahimi20203d} and attention-augmented 3D residual networks~\cite{zhang2022resattnet} report AD vs.\ CN accuracies around 90\% on ADNI-1, while hybrid CNN--Transformer models such as Conv-Swinformer~\cite{hu2023convswin} use attention to relate regions or slices. Performance on MCI vs.\ CN remains considerably lower. Wen et al.~\cite{wen2020reproducible} documented how data leakage, for example splitting at the slice rather than the subject level, inflates reported accuracies, and argued for reproducible subject-level evaluation.

\paragraph{Multimodal and contrastive fusion.} DAFT~\cite{polsterl2021daft} conditions 3D CNN feature maps on tabular data. Hager et al.~\cite{hager2023best} pretrain image and tabular encoders jointly with a contrastive objective. Huang~\cite{huang2023multimodal} adapts CLIP~\cite{radford2021clip} to ADNI by contrasting MR images with groups of tabular variables (biomarkers, cognitive tests, volumes, medical history) and adds a tabular attention module; the fused model reports 83.8\% three-way accuracy and 95.5\% on AD vs.\ CN. Other multimodal works combine imaging with clinical data for conversion or progression prediction~\cite{velazquez2022ensemble,rahim20233dcnn}.

\paragraph{Anatomical priors and explainability.} FreeSurfer~\cite{fischl2012freesurfer} and its deep-learning reimplementation FastSurfer~\cite{henschel2020fastsurfer} provide subcortical segmentation and cortical parcellation according to the Desikan--Killiany--Tourville (DKT) protocol~\cite{klein2012dkt}. Grad-CAM~\cite{selvaraju2017gradcam} is widely used to inspect which regions drive MRI classifiers~\cite{zhang2022resattnet,rahim20233dcnn}. Our work combines both: segmentation is first used to audit the classifier and then to constrain its input.

\section{Data}
\label{sec:data}

\paragraph{Imaging.} We use 1{,}075 baseline T1-weighted scans from the ADNI-1 Screening 1.5\,T collection acquired from 982 subjects; the excess of scans over subjects is due to a small number of subjects with more than one screening acquisition. Scans are labelled CN, MCI or AD, with per-class counts of approximately CN: 289 / MCI: 507 / AD: 279. All volumes were processed with FastSurfer~\cite{henschel2020fastsurfer}, which conforms each scan to a $256^3$ grid of $1\,\text{mm}^3$ isotropic voxels (\texttt{orig.mgz}) and produces a DKT~\cite{klein2012dkt} cortical parcellation with subcortical labels (\texttt{aparc.DKTatlas+aseg.deep.mgz}). Conformed volumes are stored in LIA orientation, so voxel axes 0, 1 and 2 correspond to the sagittal, axial and coronal directions, respectively. Training-time augmentation consists of random left--right flips, small random affine transforms and random bias fields.

\paragraph{Splits.} Splits are made at the subject level: no subject appears in more than one split. The test set is balanced with 21 scans per class (63 scans); binary tasks use the corresponding 42-scan subsets. The remaining 1{,}012 scans are divided into approximately 862 training and 150 validation scans.

\paragraph{Tabular data.} Clinical variables come from the consolidated ADNIMERGE table (16{,}421 subject--visit rows across all ADNI phases). Each scan is matched to a row by restricting to its subject identifier, preferring baseline visit codes (\texttt{bl}, \texttt{sc}, \texttt{scmri}) and choosing the row whose examination date is closest to the acquisition date; all 1{,}075 scans were matched. Following~\cite{huang2023multimodal}, variables are grouped into five modality groups (\cref{tab:leakage}). Features are standardized with training-set statistics; missing values in the \cog\ and \vol\ groups (fewer than 3\% of cells in either group at baseline) are imputed with the training-set mean of the corresponding feature before standardization.

\begin{table}[t]
\centering
\small
\caption{ADNIMERGE modality groups ordered by their overlap with the diagnostic label. Cognitive scores are part of the ADNI diagnostic criteria~\cite{petersen2010adni}; volumes are correlates of atrophy but not criteria.}
\label{tab:leakage}
\begin{tabular}{l p{5.6cm} p{5.2cm}}
\toprule
\textbf{Group} & \textbf{Features} & \textbf{Leakage risk} \\
\midrule
\cog  & CDR-SB, ADAS11, MMSE, RAVLT-immediate & High: diagnostic instruments \\
\bio  & A$\beta$, tau, p-tau (CSF) & Medium: biological correlates of pathology \\
\pet  & FDG, AV45 & Medium: physiological markers \\
\vol  & Hippocampus, whole brain, entorhinal, middle temporal & Low: atrophy correlates \\
\demo & APOE4, age & Low: risk factors \\
\bottomrule
\end{tabular}
\end{table}

In our experiments we use \cog\ and \vol. CSF biomarkers are available for only about 14\% of ADNIMERGE rows, so the \bio\ group would require heavy imputation, which we expect to distort contrastive alignment. The \cog\ and \vol\ groups sit at the two ends of the leakage spectrum and therefore provide the most informative contrast.

\section{Image Encoder}
\label{sec:encoder}

\begin{figure}[t]
\centering
\includegraphics[width=0.95\textwidth]{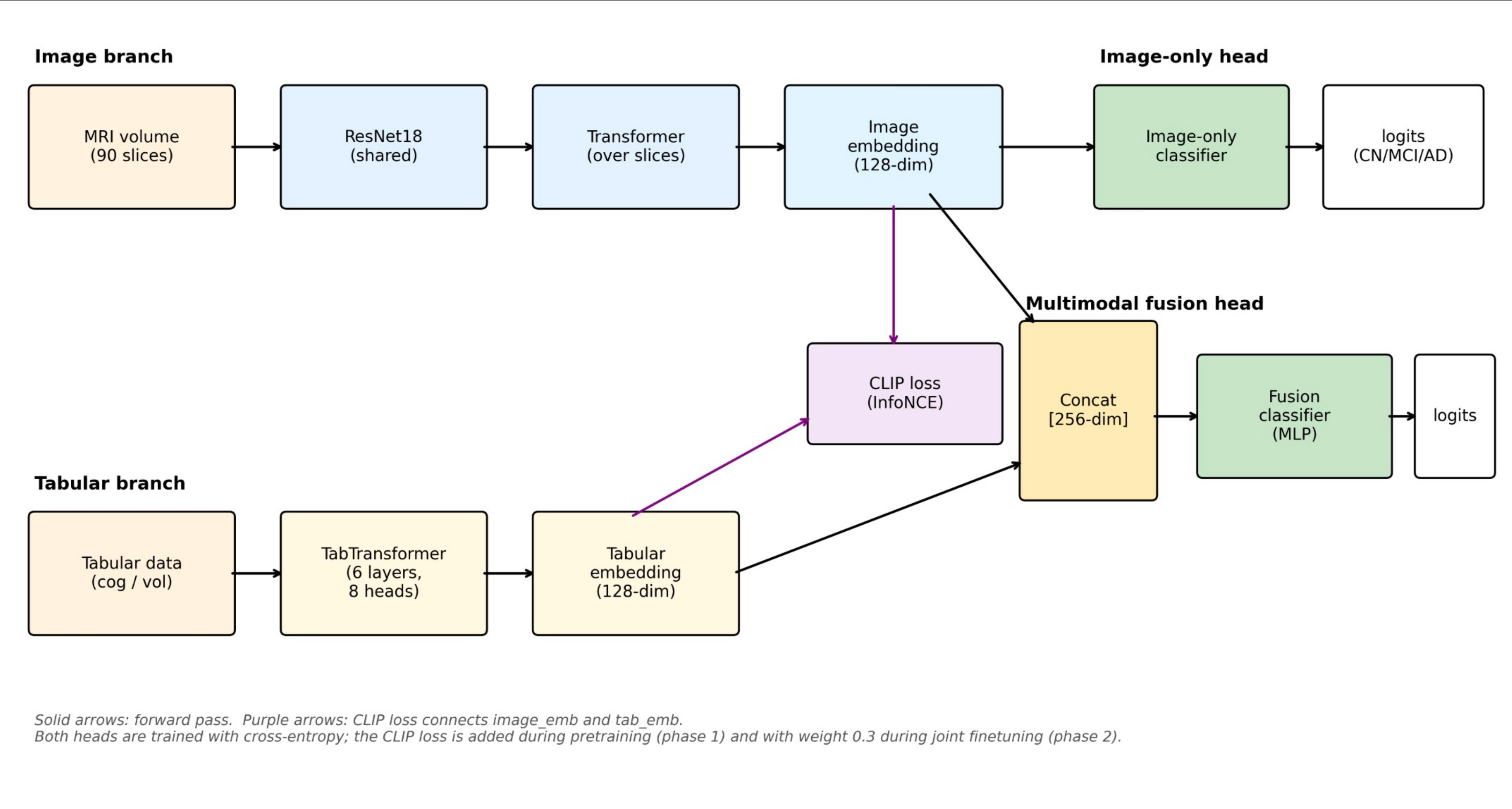}
\caption{Multimodal architecture. Image and tabular embeddings (128-d) are aligned by a CLIP loss. The image-only head classifies from the image embedding alone; the fusion head classifies from the concatenated embeddings.}
\label{fig:arch}
\end{figure}

The image branch (\cref{fig:arch}, top) processes a stack of $S$ contiguous 2D slices of a subject. Each slice is resized to $224\times224$ and passed through an ImageNet-pretrained ResNet18~\cite{he2016deep} shared across slices; only the last residual stage (\texttt{layer4}) is fine-tuned. The resulting 512-dimensional slice features are linearly projected, summed with a learned positional encoding, and processed by a single Transformer encoder layer~\cite{vaswani2017attention} with 8 attention heads. Mean pooling over slices gives a 512-dimensional subject descriptor, which a two-layer MLP maps to a 128-dimensional embedding. Dropout of 0.15 is applied in attention, feed-forward and classifier layers.

For the full-slice (``raw'') configuration we use $S=90$ slices (indices 62--152) along voxel axis~1, i.e.\ axial planes. The window was chosen with the slice statistics of \cref{sec:slices} to cover as many AD-relevant structures as possible while excluding the orbits and neck.

We chose this 2D design over 3D alternatives even though, in a preliminary comparison on an earlier split, 3D encoders reached noticeably higher accuracy (a ResNet18 3D backbone with a slice-Transformer aggregator was about 20 points ahead of the 2D counterpart). Slice-wise processing makes it straightforward to compare per-slice Grad-CAM maps with per-slice detections (\cref{sec:gradcam}), to crop anatomically (\cref{sec:mtl}) and to combine the encoder with the fusion architecture (\cref{sec:multimodal}), while keeping the parameter count low.

\section{Anatomical Reference from FastSurfer Segmentations}
\label{sec:anatomy}

\subsection{Structures and slice statistics}
\label{sec:slices}

We selected structures that are routinely inspected in AD: the hippocampus, entorhinal cortex, amygdala and parahippocampal gyrus of the MTL~\cite{pennanen2004hippocampus,dickerson2009cortical,echavarri2011parahippocampal}; the middle and inferior temporal and fusiform gyri; inferior and superior parietal cortex, precuneus, posterior and isthmus cingulate~\cite{jacobs2012parietal}; superior and rostral middle frontal gyri; the lateral and third ventricles; and cerebellar grey and white matter, all taken bilaterally. For each structure and voxel axis we counted, over 50 subjects, in how many subjects the structure's labels appear on each slice, and defined its slice range as the interval covering 90\% of occurrences. These ranges were used to choose the input window of the encoder and to decide which structures to annotate on each slice.

\subsection{Detection and segmentation with YOLOv8}

Segmentation maps were converted into YOLOv8~\cite{jocher2023yolov8} training data. For detection, each structure's binary mask on a slice was split into connected components; components smaller than 100 pixels (5 pixels for the amygdala and entorhinal cortex) were discarded, and the remaining ones were converted to bounding boxes. For instance segmentation, external contours were extracted, simplified with the Douglas--Peucker algorithm, filtered below 50 pixels, and capped at 100 polygon vertices. Slices were intensity-normalized and resized to $640\times640$. Models were trained on slices along the same axis as the encoder input (21 structures for detection, 20 for segmentation). \Cref{tab:yolo} shows that all variants localize structures reliably (mAP$_{50}\geq0.95$); the stricter mAP$_{50\text{--}95}$ still grows with model size and had not saturated within the training budget.

\begin{table}[t]
\centering
\small
\caption{YOLOv8 detection (40 epochs) and instance segmentation (15 epochs) of AD-relevant structures on held-out slices.}
\label{tab:yolo}
\begin{tabular}{llcccc}
\toprule
\textbf{Task} & \textbf{Model} & \textbf{Precision} & \textbf{Recall} & \textbf{mAP$_{50}$} & \textbf{mAP$_{50\text{--}95}$} \\
\midrule
\multirow{3}{*}{Detection}    & YOLOv8n     & 0.935 & 0.917 & 0.951 & 0.764 \\
                              & YOLOv8s     & 0.951 & 0.931 & 0.964 & 0.809 \\
                              & YOLOv8m     & 0.960 & 0.939 & 0.969 & 0.838 \\
\midrule
\multirow{3}{*}{Segmentation} & YOLOv8s-seg & 0.939 & 0.923 & 0.962 & 0.639 \\
                              & YOLOv8m-seg & 0.945 & 0.931 & 0.968 & 0.654 \\
                              & YOLOv8l-seg & 0.944 & 0.934 & 0.969 & 0.663 \\
\bottomrule
\end{tabular}
\end{table}

\subsection{What does the image classifier look at?}
\label{sec:gradcam}

\begin{figure}[t]
\centering
\begin{subfigure}[t]{0.36\textwidth}
  \centering
  \includegraphics[width=\linewidth]{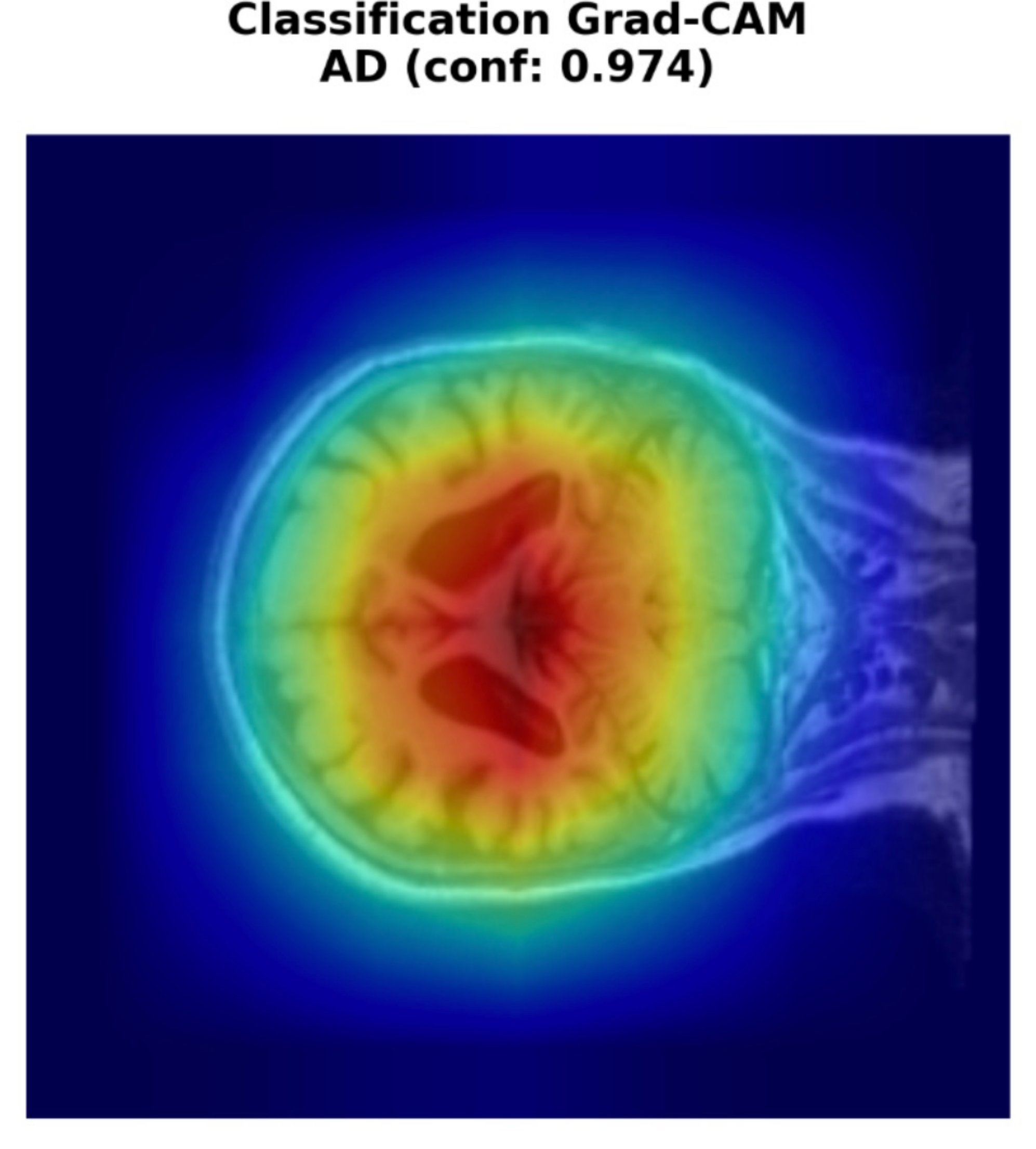}
  \caption{Slice at the level of the lateral ventricles: attention on the ventricular region.}
\end{subfigure}\hspace{0.05\textwidth}%
\begin{subfigure}[t]{0.36\textwidth}
  \centering
  \includegraphics[width=\linewidth]{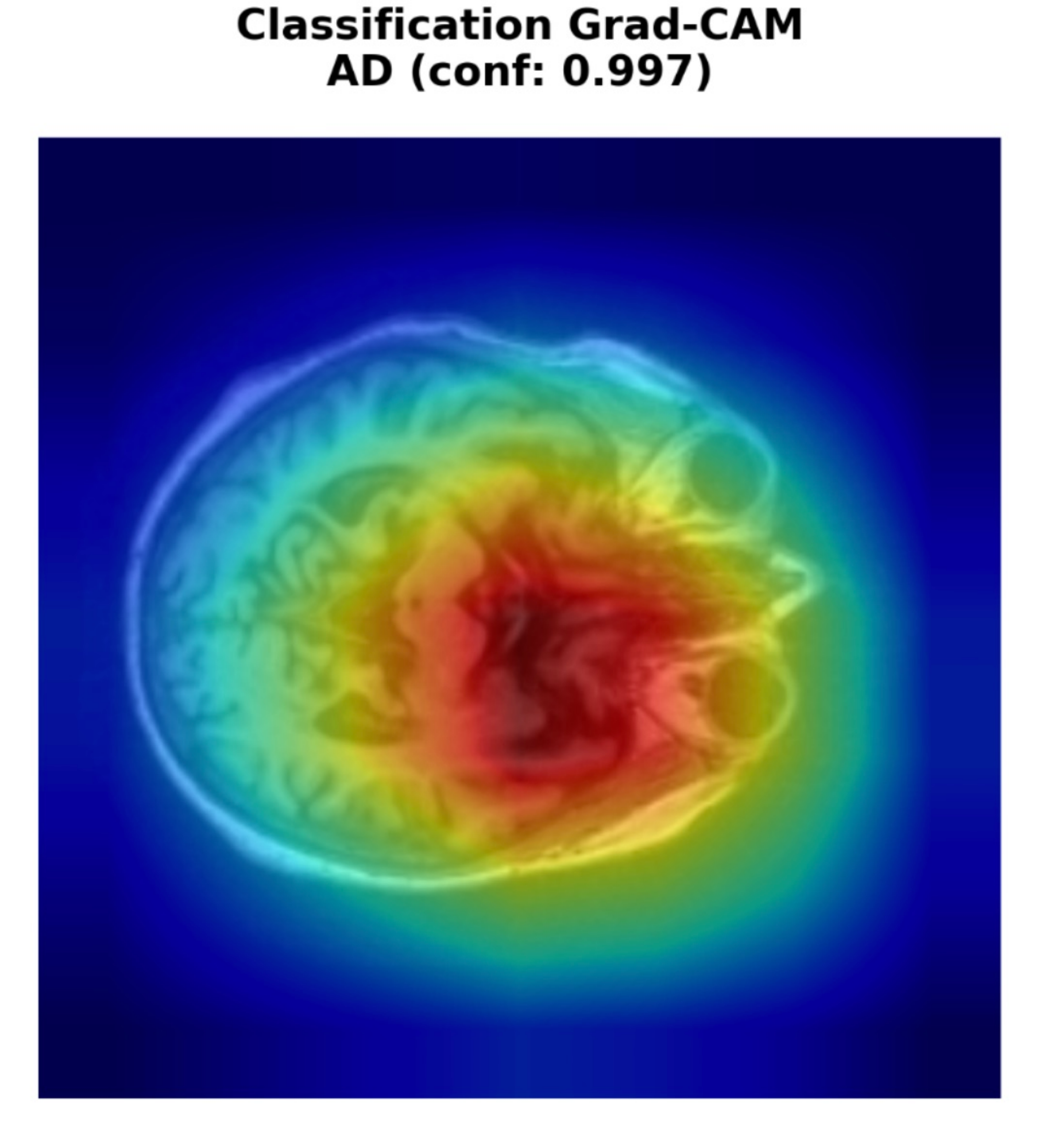}
  \caption{Slice at the level of the orbits: attention spreads to non-brain regions.}
\end{subfigure}
\caption{Grad-CAM of the raw-slice ResNet18 backbone on two input slices of AD subjects (predicted class and confidence shown above each map).}
\label{fig:gradcam}
\end{figure}

We applied Grad-CAM~\cite{selvaraju2017gradcam} to the ResNet18 backbone of the raw-slice classifier and inspected the maps slice by slice next to the YOLO detections of the same subjects. Two patterns were consistent across subjects (\cref{fig:gradcam}). On many slices, the network does attend to diagnostically meaningful tissue: when the lateral ventricles appear, attention concentrates on the high-contrast ventricular region, and on lower slices it moves to the temporal lobes, overlapping detected hippocampus, parahippocampal, fusiform, middle and inferior temporal structures. On a substantial number of other slices, however, attention falls on the skull, neck, orbits or outside the head. Correct predictions can therefore be supported by anatomically irrelevant evidence, which motivates constraining the input (\cref{sec:mtl}). This analysis is qualitative; a quantitative measure such as the fraction of Grad-CAM mass inside the brain or MTL mask is left for future work.

\section{Leakage-Aware Multimodal Contrastive Learning}
\label{sec:multimodal}

\subsection{Architecture}

\Cref{fig:arch} shows the multimodal model. The image branch is the encoder of \cref{sec:encoder}, unchanged. The tabular branch is a TabTransformer-style encoder~\cite{huang2020tabtransformer}: each of the $F$ features of a modality group ($F=4$ for \cog\ and \vol) is linearly tokenized into a 32-dimensional embedding with a learned positional embedding; the token sequence passes through a 6-layer Transformer with 8 heads of dimension 16; tokens are flattened, layer-normalized and projected to 128 dimensions.

Let $z^{I}_i$ and $z^{T}_i$ be the $\ell_2$-normalized image and tabular embeddings of subject $i$ in a batch of size $N$, and $s_{ij}=\exp(t)\,z^{I\top}_i z^{T}_j$ with learnable temperature $t$ initialized at $\log(1/0.07)$. The symmetric CLIP loss~\cite{radford2021clip} is
\begin{equation}
\mathcal{L}_{\text{CLIP}} = -\frac{1}{2N}\sum_{i=1}^{N}\left[\log\frac{e^{s_{ii}}}{\sum_{j} e^{s_{ij}}} + \log\frac{e^{s_{ii}}}{\sum_{j} e^{s_{ji}}}\right].
\end{equation}
We compute this loss in fp32: under mixed precision it remained at its chance value $\log N$.

Two heads are trained on top of the embeddings. The \emph{image-only head} is a linear classifier on $z^{I}$ and measures what the image encoder has learned by itself; it uses no tabular input at inference. The \emph{fusion head} is a two-layer MLP with dropout on the 256-dimensional concatenation $[z^{I}; z^{T}]$.

\subsection{Training}

When cross-entropy and contrastive terms were optimized jointly from the start, the cross-entropy gradients dominated and the CLIP loss stayed near $\log N$. We therefore train in two phases: (i) contrastive pretraining with $\mathcal{L}_{\text{CLIP}}$ only for 8--10 epochs; (ii) joint fine-tuning for 20--25 epochs with
\begin{equation}
\mathcal{L} = \mathcal{L}_{\text{CE}}^{\text{img}} + \mathcal{L}_{\text{CE}}^{\text{fus}} + 0.3\,\mathcal{L}_{\text{CLIP}}.
\end{equation}
We use AdamW~\cite{loshchilov2019adamw} with a cosine learning-rate schedule, gradient clipping, class-balanced sampling, label smoothing of 0.1, and an exponential moving average of the weights for evaluation. Each modality group is trained in a separate run.

\subsection{Why the leakage spectrum matters}

In ADNI-1, CN, MCI and AD participants were defined by MMSE ranges, the global Clinical Dementia Rating and memory testing, together with clinical judgement~\cite{petersen2010adni}. The \cog\ group contains the MMSE and the CDR sum of boxes, which is closely related to the global CDR. Predicting the label from these variables therefore largely amounts to re-deriving the rule that produced it. We report \cog\ fusion as an upper bound and a sanity check of our implementation, and treat \vol\ fusion, whose features are atrophy measurements rather than diagnostic criteria, as the imaging-relevant result.

\section{Results}
\label{sec:results}

All accuracies are computed on the balanced subject-level test set: 63 scans for three-way classification and 42 for each binary task. Brackets give 95\% Wilson confidence intervals~\cite{wilson1927probable}; on 63 scans the interval spans roughly $\pm 12$ percentage points. All results are single runs.

\subsection{Three-way classification}

\begin{table}[t]
\centering
\small
\caption{Three-way (CN / MCI / AD) test accuracy, 63 scans (21 per class). The last two rows are reported by Huang~\cite{huang2023multimodal} under a different protocol (882 test slices, all tabular groups fused, mean $\pm$ s.d.\ over 5 splits) and are listed for reference only.}
\label{tab:3way}
\begin{tabular}{lcc}
\toprule
\textbf{Configuration} & \textbf{Correct} & \textbf{Accuracy (\%)} \\
\midrule
Image only (raw slices)                  & 37/63 & 58.7\ci{46.4}{70.0} \\
Multimodal, \cog\ (upper bound, leakage) & 55/63 & 87.3\ci{76.9}{93.4} \\
Multimodal, \vol                          & 46/63 & 73.0\ci{61.0}{82.4} \\
\midrule
Huang~\cite{huang2023multimodal}, MR images only        & -- & 76.1 $\pm$ 1.4 \\
Huang~\cite{huang2023multimodal}, multimodal (all data) & -- & 83.8 $\pm$ 2.3 \\
\bottomrule
\end{tabular}
\end{table}

\Cref{tab:3way} shows that fusion with cognitive scores raises accuracy from 58.7\% to 87.3\%, with balanced per-class performance (19/21 MCI, 17/21 CN, 19/21 AD). This confirms that the contrastive pipeline is functioning, but for the reasons in \cref{sec:multimodal} it is not evidence of better image understanding. Fusion with volumes also improves over the image-only model (73.0\%); its residual errors are concentrated in the MCI/CN distinction, where baseline hippocampal and entorhinal volumes of the two groups overlap substantially at the subject level.

\subsection{Binary tasks}

\begin{table}[t]
\centering
\small
\caption{Binary test accuracy (\%), 42 scans per task (21 per class). ``Image head (vol-aligned)'' is the image-only head of the model trained jointly with the \vol\ contrastive target; ``Image head (cog-aligned)'' is the image-only head of the model trained with the \cog\ target. Reference rows come from other protocols and are not directly comparable.}
\label{tab:binary}
\begin{tabular}{lcc}
\toprule
\textbf{Configuration} & \textbf{AD vs.\ CN} & \textbf{MCI vs.\ CN} \\
\midrule
Image head (vol-aligned)                  & 85.7\ci{72.2}{93.3} & 73.8\ci{58.9}{84.7} \\
Image head (cog-aligned)                  & -- & 52.4\ci{37.7}{66.6} \\
Multimodal, \cog\ (upper bound, leakage)  & 95.2\ci{84.2}{98.7} & 83.3\ci{69.4}{91.7} \\
Multimodal, \vol                          & 83.3\ci{69.4}{91.7} & 78.6\ci{64.1}{88.3} \\
\midrule
Huang~\cite{huang2023multimodal}, MR images only          & 88.5 $\pm$ 1.5 & -- \\
Huang~\cite{huang2023multimodal}, multimodal (all data)   & 95.5 $\pm$ 1.7 & -- \\
Zhang et al.~\cite{zhang2022resattnet}, 3D ResAttNet (MRI, ADNI-1) & 91.3 $\pm$ 1.2 & -- \\
\bottomrule
\end{tabular}
\end{table}

On AD vs.\ CN, \cog\ fusion misclassifies only two of 42 scans (CN 21/21, AD 19/21). The vol-aligned image head reaches 85.7\%, which illustrates that gross late-stage atrophy is largely visible to the image encoder; \vol\ fusion does not improve over the image head on this easier task. On MCI vs.\ CN, both fusion variants improve over the image head, and the vol-aligned head substantially outperforms the cog-aligned head (\cref{sec:target}). We are not aware of an MCI vs.\ CN result in~\cite{huang2023multimodal} and therefore list no reference for this task.

\section{The Contrastive Target Shapes the Image Encoder}
\label{sec:target}

The results above are accuracies of models. This section reports what we consider the main finding of the study, which concerns the \emph{representation} rather than the accuracy: changing only the tabular group that the CLIP loss aligns the image embedding to changes what the image encoder itself learns.

\begin{figure}[t]
\centering
\includegraphics[width=0.55\textwidth]{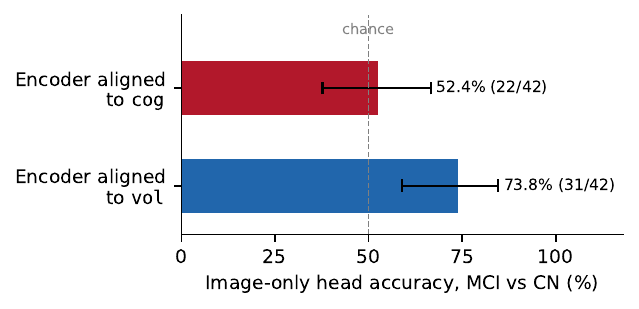}
\caption{Image-only head accuracy on MCI vs.\ CN for encoders aligned to cognitive scores (\cog) or regional volumes (\vol). Same architecture, training schedule, split and test set (42 scans); error bars are 95\% Wilson intervals.}
\label{fig:target}
\end{figure}

\paragraph{A controlled comparison.} The image-only head takes no tabular input at inference, so its accuracy measures the visual representation alone. We compare this head across two runs that are identical in architecture, training data, subject split, two-phase schedule, hyperparameters and test set. The single difference is the contrastive target: in one run the encoder is aligned to \cog, in the other to \vol. On MCI vs.\ CN, the \cog-aligned image head is near chance (52.4\%, 22/42), while the \vol-aligned head reaches 73.8\% (31/42), a gap of 21 points obtained without the tabular branch being present at evaluation (\cref{fig:target}).

\paragraph{Interpretation.} A plausible account is that the geometry of the target determines what the image branch must encode in order to match it. Regional volumes live in a spatial, structural feature space: an image embedding can only be pulled towards them by encoding hippocampal, entorhinal and global atrophy, and those features remain discriminative once the tabular branch is removed. Cognitive scores offer no spatial structure to latch onto. Worse, because they nearly determine the label (\cref{tab:leakage}), the fusion head can solve the task from the tabular branch alone, so neither the contrastive nor the classification gradient pushes the encoder to extract atrophy-related features. The fused accuracies invert the ordering of the two runs, 87.3\% for \cog\ against 73.0\% for \vol\ on the three-way task, which is exactly why the image-only head has to be reported alongside the fused one: the configuration that looks better end-to-end is the one that learned the weaker visual representation.

\paragraph{Consequences.} Two practical points follow. For benchmarking, a fused number alone cannot tell whether a multimodal model improved its use of imaging or merely gained access to the labelling rule; reporting the unimodal head under each target separates the two. For training, the tabular side can be treated as a design choice that shapes the encoder rather than as extra input, which is the reasoning that led us to restrict the input to the same anatomical region in \cref{sec:mtl}.

\paragraph{Strength of evidence.} The gap corresponds to nine test scans. An unpaired two-proportion $z$-test gives $p\approx0.04$ for a single run, so we regard the effect as suggestive rather than established, and its size should not be taken literally. Confirming it requires a paired test on per-scan predictions, repetition over several seeds, and ideally a third target of intermediate leakage such as \pet\ or \demo.

\section{Anatomically Guided Input: MTL Crop}
\label{sec:mtl}

\begin{figure}[t]
\centering
\includegraphics[width=\textwidth]{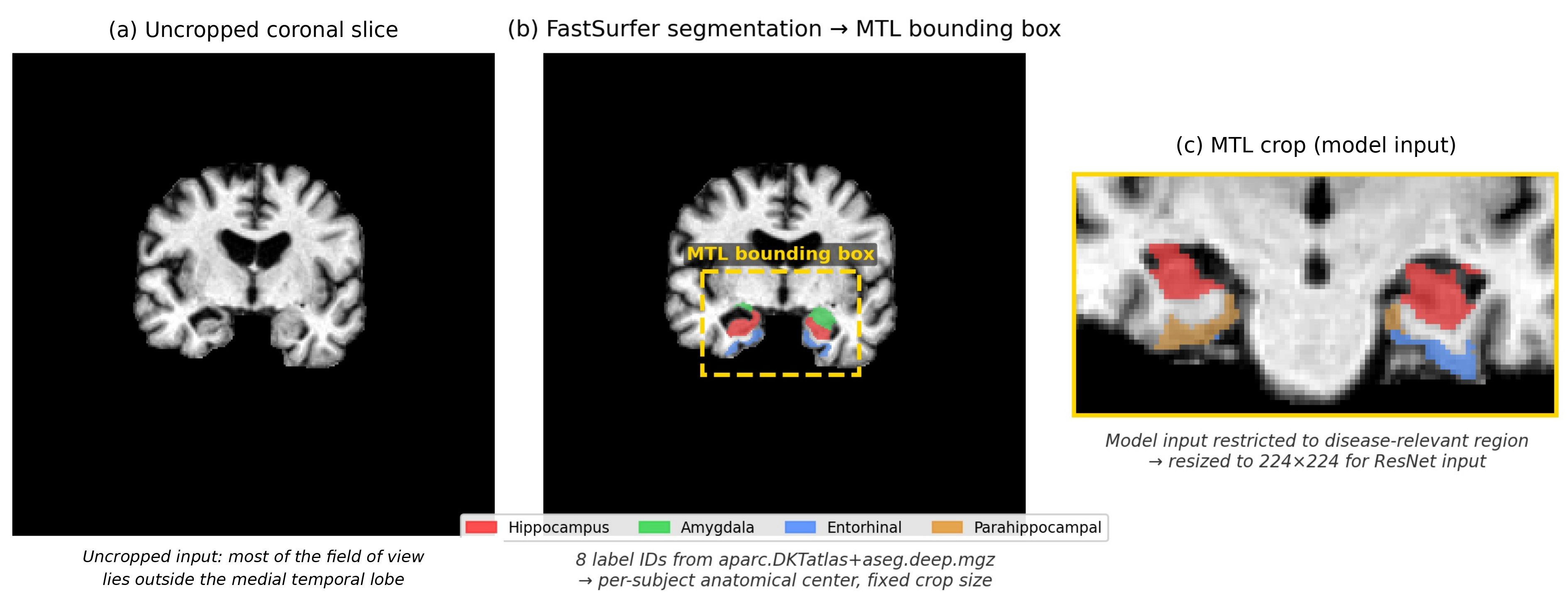}
\caption{MTL crop for an AD subject. (a) Uncropped coronal slice. (b) FastSurfer labels of the eight MTL structures and the bounding box of their union, whose centroid defines the crop centre. (c) The fixed-size crop given to the encoder after resizing to $224\times224$.}
\label{fig:mtl}
\end{figure}

Motivated by the Grad-CAM analysis, we restrict the encoder input to the medial temporal lobe. For each subject, we form a binary mask from the union of eight FastSurfer labels, bilateral hippocampus (17, 53), amygdala (18, 54), entorhinal cortex (1006, 2006) and parahippocampal cortex (1016, 2016), and compute its centroid. A fixed-size box of $96\times48\times64$ voxels (left--right $\times$ inferior--superior $\times$ anterior--posterior) centered on this subject-specific centroid is extracted, so input size is constant while position follows each subject's anatomy (\cref{fig:mtl}). The crop is sliced along the coronal axis (voxel axis~2, 64 slices of $96\times48$), each slice resized to $224\times224$, and passed to the unchanged encoder. The MTL was chosen because it carries the earliest structural changes in AD~\cite{pennanen2004hippocampus,dickerson2009cortical,echavarri2011parahippocampal} and overlaps with the \vol\ features, so the input and the contrastive target emphasize the same region.

\begin{table}[t]
\centering
\small
\caption{Raw-slice vs.\ coronal MTL-crop input. Three-way on 63 scans, binary on 42 scans. The ``Image only, raw slices'' binary numbers are the image head of the vol-aligned jointly trained model of \cref{tab:binary}; for a fair comparison the ``Image only, MTL crop'' rows also report the image head of a vol-aligned jointly trained model on the cropped input, rather than a cross-entropy-only model.}
\label{tab:mtl}
\begin{tabular}{lccc}
\toprule
\textbf{Configuration} & \textbf{3-way} & \textbf{AD vs.\ CN} & \textbf{MCI vs.\ CN} \\
\midrule
Image only, raw slices              & 58.7 & 85.7 & 73.8 \\
Image only, MTL crop (coronal)       & 65.1 & 88.1 & 81.0 \\
\midrule
Multimodal \vol, raw slices          & 73.0 & 83.3 & 78.6 \\
Multimodal \vol, MTL crop (coronal)  & 76.2 & 90.5 & 85.7 \\
\bottomrule
\end{tabular}
\end{table}

\Cref{tab:mtl} summarizes the results. With the coronal crop, image-only three-way accuracy increases from 58.7\% (37/63) to 65.1\% (41/63, 95\% CI [52.8, 75.7]); the MCI vs.\ CN task benefits more than AD vs.\ CN, consistent with the MTL carrying the early, subtle signal while late-stage atrophy is already visible in full slices. Coronal is the standard view for the hippocampus and entorhinal cortex; we also trained axial and sagittal variants of the crop, but the numbers are omitted here pending final log extraction. Combining the crop with \vol\ alignment helps further but sub-additively, which we attribute to both interventions targeting the same region. We stress that the three-way improvement corresponds to four scans and its confidence interval overlaps with the baseline.

\section{Discussion and Limitations}
\label{sec:discussion}

\paragraph{Comparison with published results.} Our \cog\ numbers are numerically close to those of Huang~\cite{huang2023multimodal}, but the protocols differ in ways that preclude a ranking: (i) we evaluate 63 scans at the subject level, whereas~\cite{huang2023multimodal} evaluates 882 slices; (ii) we use ADNI-1 baseline scans only, whereas~\cite{huang2023multimodal} uses a different and larger selection of ADNI images; (iii) we fuse one modality group at a time, whereas~\cite{huang2023multimodal} fuses all groups, including a medical-history group that contains the baseline diagnosis; (iv) we report single runs, whereas~\cite{huang2023multimodal} averages five splits. We read the proximity of the numbers as evidence that our re-implementation works, not as an improvement. Our image-only baselines remain below 3D encoders on ADNI~\cite{huang2023multimodal,zhang2022resattnet}; the MTL crop narrows but does not close this gap.

\paragraph{Leakage.} Cognitive scores are legitimate clinical inputs, and a deployed system may well use them. For assessing what an imaging model contributes, however, they conflate image information with the labelling rule. Our results suggest a further effect: a leaky contrastive target can also \emph{degrade} the image representation, since the image-only head of the \cog-aligned model is near chance on MCI vs.\ CN. We recommend reporting fusion results per modality group and always reporting the image-only head alongside the fused one.

\paragraph{Limitations.} The test set is small and every configuration is a single run, so differences of a few points are within noise. The MTL crop requires a FastSurfer segmentation at inference, and segmentation errors propagate directly to the crop position. All data come from one cohort and field strength (ADNI-1, 1.5\,T), and Grad-CAM findings are qualitative. Our compute environment imposed a 12-hour limit per job, which prevented longitudinal multi-phase training.

\paragraph{Future work.} We plan to (i) repeat all experiments over several seeds and a larger, longitudinal ADNI-1/2/3 set with cached slice features; (ii) add the \bio\ group where CSF coverage permits and test whether it induces a similar effect on the image encoder; (iii) quantify Grad-CAM overlap with segmented structures and analyze disagreements between explanation and prediction; and (iv) reduce the dependence on external segmentation, for example through semi-supervised training.

\section{Conclusion}

Using a lightweight slice-based encoder on ADNI-1, we showed that anatomical references derived from FastSurfer can be used both to audit and to constrain an Alzheimer's classifier, that multimodal gains obtained with cognitive scores largely reflect label leakage, and that the choice of contrastive target influences what the image encoder learns, with volume alignment yielding a markedly better image-only representation for MCI vs.\ CN than alignment to cognitive scores. Restricting the input to the medial temporal lobe further improves image-only accuracy. Given the small test set and single runs, these findings should be confirmed with repeated, larger-scale experiments.

\section*{Acknowledgements}
Data collection and sharing for this project was funded by the Alzheimer's Disease Neuroimaging Initiative (ADNI) (National Institutes of Health Grant U01 AG024904) and DOD ADNI (Department of Defense award number W81XWH-12-2-0012). Data used in preparation of this article were obtained from the Alzheimer's Disease Neuroimaging Initiative (ADNI) database (adni.loni.usc.edu). As such, the investigators within the ADNI contributed to the design and implementation of ADNI and/or provided data but did not participate in analysis or writing of this report. A complete listing of ADNI investigators can be found at: \url{http://adni.loni.usc.edu/wp-content/uploads/how_to_apply/ADNI_Acknowledgement_List.pdf}. Experiments were run on the FEP cluster of the National University of Science and Technology POLITEHNICA Bucharest.

\end{document}